\pdfoutput=1

\documentclass[11pt]{article}

\usepackage[final]{acl}

\usepackage{times}
\usepackage{latexsym}

\usepackage[T1]{fontenc}

\usepackage[utf8]{inputenc}

\usepackage{microtype}

\usepackage{inconsolata}

\usepackage{graphicx}
\usepackage{booktabs}
\usepackage{amssymb}

\title{Visual Prompting in Multimodal Large Language Models: A Survey}

\author{%
Junda Wu$^1$ \quad 
Zhehao Zhang$^2$ \quad
Yu Xia$^1$ \quad
Xintong Li$^1$ \quad 
Zhaoyang Xia$^3$ \quad 
Aaron Chang$^4$ \\
\textbf{Tong Yu}$^5$ \quad 
\textbf{Sungchul Kim}$^5$ \quad 
\textbf{Ryan A. Rossi}$^5$ \quad 
\textbf{Ruiyi Zhang}$^5$ \quad
\textbf{Subrata Mitra}$^5$  \\
\textbf{Dimitris N. Metaxas}$^3$ \quad
\textbf{Lina Yao}$^{6,7}$ \quad 
\textbf{Jingbo Shang}$^{1}$ \quad 
\textbf{Julian McAuley}$^{1}$ \quad \\ 
$^1$UC San Diego \quad 
$^2$Dartmouth College \quad 
$^3$Rutgers University \quad
$^4$UC Los Angeles    \\
$^5$Adobe Research \quad 
$^6$The University of New South Wales \quad
$^7$CSIRO's Data61  \\
\texttt{\{juw069,yux078,xil240,jshang,jmcauley\}@ucsd.edu} \quad
\texttt{zhehao.zhang.gr@dartmouth.edu} \\
\texttt{zx149@rutgers.edu} \quad
\texttt{aaronchang21@ucla.edu} \quad
\texttt{dnm@cs.rutgers.edu} \\
\texttt{\{tyu,sukim,ryrossi,ruizhang,sumitra\}@adobe.com} \quad
\texttt{lina.yao@data61.csiro.au} \\
}

\begin{document}
\maketitle

\begin{abstract}
Multimodal large language models (MLLMs) equip pre-trained large-language models (LLMs) with visual capabilities.
While textual prompting in LLMs has been widely studied,
visual prompting has emerged for more fine-grained and free-form visual instructions.
This paper presents the first comprehensive survey on visual prompting methods in MLLMs,
focusing on visual prompting, prompt generation, compositional reasoning, and prompt learning.
We categorize existing visual prompts and discuss generative methods for automatic prompt annotations on the images. 
We also examine visual prompting methods that enable better alignment between visual encoders and backbone LLMs,
concerning MLLM's visual grounding, object referring, and compositional reasoning abilities.
In addition, we provide a summary of model training and in-context learning methods to improve MLLM's perception and understanding of visual prompts.
This paper examines visual prompting methods developed in MLLMs and provides a vision of the future of these methods.
\end{abstract}

\section{Introduction}
Multimodal large language models (MLLMs) \cite{li2023blip,liu2024visual}, 
which augment pre-trained large language models (LLMs) with visual capabilities, 
enable visual understanding and reasoning on complex multimodal tasks \cite{zhou2024image,jia2024describe}.
However, limited by using textual prompts to describe and specify visual elements \cite{lin2024draw,wu2024controlmllm},
conventional prompting methods fall short of providing accurate visual grounding and referring to detailed visual information,
which can cause visual hallucinations \cite{bai2024hallucination,huang2024visual} and language bias \cite{wu2024commit,qu2024unified}.

Recently, visual prompting methods have emerged \cite{zhang2024vpgtrans,wu2024dettoolchain} as a new paradigm, 
complementing textual prompting and enabling more fine-grained and pixel-level instructions on multimodal input.
Since visual prompting methods can take heterogeneous forms for various tasks and often operate at pixel-level granularity, 
general prompt templates might not apply to different images, making instance-level visual prompt generation necessary.
Therefore, we provide a comprehensive categorization of current visual prompting methods (Section \ref{sec:type}) 
and methods to generate (Section \ref{sec:gen}) such visual prompts.

Despite the success of visual prompting methods in augmenting MLLM's visual abilities,
several works also suggest that MLLMs can be misaligned with visual prompts, 
due to the lack of heterogeneous visual prompting training data during the pre-training stage \cite{yan2024list,lin2024rethinking}.
This misalignment can cause MLLMs to neglect or misinterpret certain visual prompts, causing hallucination problems.
Therefore, we summarize existing efforts in aligning visual prompting with MLLM's perception and reasoning 
enabling more controllable compositional reasoning (Section \ref{sec:reason}).
In addition, we examine existing pre-training, fine-tuning (Section \ref{sec:train}), and in-context learning methods (Section \ref{sec:icl}) 
that fundamentally align MLLMs with multimodal augmented prompts. 

\begin{figure*}
    \centering
    \includegraphics[width=1\textwidth]{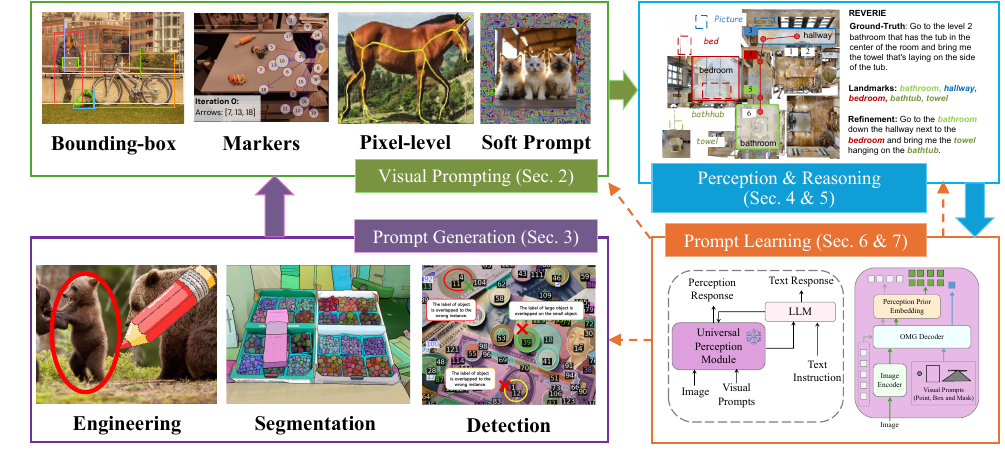}
    \caption{
    Taxonomy flow chart of visual prompting techniques. 
    We illustrate in order of four stages of visual prompting including prompt generation, visual prompting, perception and reasoning, and prompt learning, 
    where the solid arrows show the direction of each component's information flow.
    We explain in detail various visual prompt generation techniques (Section \ref{sec:gen}), and how these generated prompts are used to prompt MLLMs (Section \ref{sec:type}). 
    Then we discuss the advanced perception and reasoning abilities achieved through visual prompting (Section \ref{sec:perception} and \ref{sec:reason}). 
    Finally, model pre-training, fine-tuning, instruction tuning, and in-context learning further update previous model components, which are illustrated by the dashed arrows (Section \ref{sec:train} and \ref{sec:icl}).
    }
    \label{fig:overview}
    \vspace{-1.em}
\end{figure*}

Existing surveys on LLM prompting are limited to textual prompt design \cite{gu2023systematic, schulhoff2024prompt} and in-context demonstrations \cite{xu2024context, li2023practical}, which lack literature coverage of pixel-level instructions and multimodal interactions. 
Visual prompting is also studied in computer vision. 
However, relevant surveys are limited to vision tasks with vision backbone models \cite{lei2024prompt, zhang2024vision}, 
while multimodal perception and reasoning tasks involving MLLMs are absent.
In addition, one recent survey on Segment Anything Models (SAM) \cite{zhang2023survey} explores various applications of SAM in MLLMs.
However, this work is limited to the SAM model and lacks comprehensive studies on diverse visual prompting methods.
In this paper, we present the first comprehensive survey on visual prompting in MLLMs to address these gaps and extend the current understanding of visual prompt generation, multimodal prompting, perception and reasoning, and prompt learning.
We illustrate the taxonomy of our survey in Figure \ref{fig:overview} and summarize our contributions as follows:
\begin{itemize}
    \item We provide a comprehensive categorization of visual prompting and prompt generation methods in MLLMs.
    \item We explain the integration of visual prompts into MLLM's perception and reasoning for more controllable compositional reasoning,
    which helps to prevent hallucination and language bias issues. 
    \item We summarize MLLM alignment methods with visual prompts, covering model training and in-context learning, 
    addressing issues of misinterpretation, and proposing strategies for more controllable compositional reasoning.
\end{itemize}

\section{Visual Prompt Categorization} \label{sec:type}
Visual prompts are essential tools in MLLMs, guiding models in interpreting and processing visual data. 
These prompts~\cite{wu2024dettoolchain} can take various forms, such as bounding boxes, markers, pixel-level prompts, and soft prompts. 
They provide additional information to enhance the models' visual perception capabilities. 
By manipulating images and videos with different techniques, visual prompts improve model performance in complex understanding and reasoning tasks.

\subsection{Bounding-Box}
Bounding boxes are used to demarcate objects or regions within an image, enabling MLLMs to extract visual features~\cite{lin2024draw}.  
These features help the model understand the image content and correlate it with the corresponding text, 
thereby enhancing fine-grained and grounded image understanding. 
Previous work, such as Shikra~\cite{chen2023shikra} and VTPrompt~\cite{jiang2024joint}, quantize bounding boxes to represent key objects numerically, 
modeling both input and output positions. 
Other approaches modify bounding boxes for specific tasks: 
A3VLM~\cite{huang2024a3vlm} uses 3D bounding boxes to locate actionable parts of an image, 
CityLLaVA~\cite{duan2024cityllava} scales up the bounding box, and TextCoT~\cite{luan2024textcot} extends the shorter sides of the bounding box to match the longer side, ensuring it encompasses the entire region of interest. 
In addition, CRG~\cite{wan2024contrastive} masks out specific regions with black pixels to reduce priors, providing a way to correct predictions without additional training. 
Groma~\cite{ma2024groma} and InstructDET~\cite{dang2023instructdet} encode user-specified regions (i.e., bounding boxes) into visual tokens, enhancing the localization ability of MLLMs by directly integrating them into user instructions. 
Another framework~\cite{lin2024rethinking} further enhances the localization capabilities of MLLMs by integrating contextual embeddings from external knowledge within bounding boxes, serving as visual prompts to boost the fine-grained cognitive abilities of various MLLMs. 

\subsection{Markers}
Similar to bounding boxes, visual markers are specific elements within visual data (such as images or videos) used to highlight, identify, or draw attention to particular features or regions. 
They are often employed to indicate particular parts of an image that are relevant to the task. 
Prior work~\cite{shtedritski2023does} has demonstrated that models trained on web-scale data can focus on specific visual markers, such as red circles, 
to highlight desired regions instead of cropping the image around them. 
AutoAD-Zero\cite{xie2024autoad} introduced a two-stage, training-free approach that incorporates character information 
by "circling" characters in the frame and color-coding each identity. 
More recently, Set-of-Mark (SoM) prompting\cite{yang2023set} overlays visual markers directly onto images to help models generate answers grounded in specific image regions. 
ViP-LLaVA\cite{cai2024vip} expands on this by incorporating arbitrary visual cues like scribbles and arrows, using fine-tuned models to recognize these markers. 
\citet{liao2024can} also leverage the SoM technique to introduce feedback, converting it into text or visual marks to improve semantic grounding. 
SoM-LLaVA \cite{yan2024list} proposes a method to enhance SoM's tag association by listing items individually and comprehensively describing all tagged items within an image. 
Other methods, such as ToL~\cite{fan2024read} and OWG~\cite{tziafas2024towards}, link each segment in the frame with a unique ID, 
while Pivot~\cite{nasiriany2024pivot} projects a 3D location into image space and draws a visual marker at this projected location to refer to spatial concepts in the output space.

\subsection{Pixel-level}
Previous approaches relied on coarse markers like colorful boxes or circles, which introduced ambiguity in accurately highlighting objects. 
To address this issue, pixel-level prompts~\cite{ma2024invariant} use individual pixels in images or videos, 
enhancing the semantic localization capability of MLLMs. 
Methods such as FGVP~\cite{yang2024fine}, EVP~\cite{liu2023explicit}, DOrA~\cite{wu2024dora}, and CoLLaVO~\cite{lee2024collavo} employ pixel-level prompts to convey semantic information for precise object localization. 
OMG-LLaVA~\cite{zhang2024omg} and VisionLLM~\cite{wang2024visionllm} tokenize images into pixel-centric visual tokens, 
aligning visual tasks with language instructions. 
Techniques such as Image Inpainting~\cite{bar2022visual} decode visual tokens into pixels, 
while ControlMLLM~\cite{wu2024controlmllm} models rich semantic relations between pixels and text prompts. 
There are also coordinate prompt methods, such as SCAFFOLD~\cite{lei2024scaffolding} and AO-Planner~\cite{chen2024affordances}, 
which convert input images into coordinates using metrics, enhancing spatial understanding and reasoning abilities in MLLMs.

\subsection{Soft Visual Prompt}
Soft visual prompts, learned in the pixel space and applied directly to the image, allow models to adapt more effectively to specific downstream tasks. 
In particular, TVP~\cite{zhang2024exploring}, BlackVIP~\cite{oh2023blackvip}, and VPGTrans~\cite{zhang2024vpgtrans} add pixel-level prompts to images, 
either by surrounding the image with universal prompts or designing prompts matching the image's shape. 
In Learned Prompt~\cite{rezaei2024learning}, WVPrompt~\cite{ren2024you}, and ILM-VP~\cite{chen2023understanding}, 
task-relevant perturbation patterns are injected into the pixel space to modify the input sample. 
Additionally, ImageBrush~\cite{yang2024imagebrush} enhances semantic understanding by extracting tokenized features from images.

\section{Visual Prompt Generation }\label{sec:gen}
Different from textual prompts, visual prompts are typically position-aware and instance-specific, involving particular visual objects, relationships, and contexts. 
Current approaches use visual prompt generation methods and models to improve the accuracy and comprehension of visual prompts by MLLMs, which generate visual prompts, such as segmentation, detection, and image inpainting, for individual images and videos.
Additionally, toolchains of visual prompt methods are employed to enable multi-step visual reasoning and planning. 
To create universally applicable visual prompts 
learnable pixel values have also been developed.

\subsection{Prompt Engineering}
Understanding human-engineered visual prompts can be important in practical use cases, where visual prompts are especially efficient for expressing one's intention or attention to the current visual evidence. Early exploration \cite{shtedritski2023does} discovers that drawing a simple red circle around an object can direct a model's attention to that region. In addition, to enrich detailed visual evidence, MIVPG \cite{zhong2024enhancing} leverages instance correlations within images or patches. 

ViP \cite{cai2024vip} introduces a novel multimodal model capable of decoding free-form visual prompts, allowing users to intuitively mark images with natural cues. This approach does not require complex region encodings and achieves state-of-the-art performance on region-specific comprehension tasks. In addition, ViP-Bench \cite{cai2024vip} is also proposed to evaluate MLLM's perception of such naturally engineered visual prompts.
In domain-specific CityLLaVA \cite{duan2024cityllava} framework, engineered visual prompts are collected and tailored for urban scenarios, which further augments the fine-tuned MLLM.

\subsection{Visual Segmentation}
Segmentation methods such as OpenSeeD \cite{zhang2023simple}, SAM \cite{kirillov2023segment}, and SegFormer \cite{xie2021segformer}, are used to delineate and identify specific regions, objects, or structures within images, thus enabling the models to focus on relevant visual information more accurately.
With pre-trained segmentation models, external visual knowledge can be transferred and integrated into MLLM's prompt.
\citet{yang2024fine} explore a fine-grained visual prompting method by pixel-level annotations annotated from image inpainting \cite{bar2022visual} method. 
\citet{lin2024rethinking} propose an instruction tuning method to directly incorporate fine-grained segmentation knowledge in the spatial embedding map as visual prompts, which enhances the model's context-awareness of the visual scene.
VAP \cite{chen2024affordances} develops a visual affordance prompting method that grounds visual elements by SAM \cite{kirillov2023segment} in navigation tasks.
DOrA \cite{wu2024dora} further introduces 3D spatial and contextual information to improve 3D visual grounding tasks.

Fine-grained segmentation information also augments MLLM's visual perception and reasoning abilities. 
OMG-LLaVA \cite{zhang2024omg} integrates multi-level visual prompts that enable MLLM's course-to-fine visual perception to more comprehensive visual understanding.
\citet{liu2023explicit} propose to enhance the model's ability to understand and process low-level structural elements within images.
\citet{he2024multi} further incorporate such visual prompts into MLLM fine-tuning to augment the model's capacity in fine-grained visual perception.
CoLLaVO \cite{lee2024collavo} proposes a crayon prompting which further augments with panoptic segmentation method with image in-painting color maps to better discriminate multi-objects within the image.

\subsection{Object Detection}
Object detection models like SoM \cite{yang2023set}, RCNN \cite{girshick2015fast}, and Omni3D \cite{brazil2023omni3d} provide precise object identification and localization in the visual context, which assists MLLM's visual grounding abilities and guides MLLM's attention on semantically meaningful contents. 
SoM-LLaVA developed by \citet{yan2024list} uses numeric tags to align visual objects with textual descriptions. 
Object tags enable the model to list and describe these objects accurately, which enhances visual reasoning and visual instruction following capabilities.
InstructDET \cite{dang2023instructdet} incorporates generalized instructions into the training process, 
diversifying object detection by enabling the model to understand and follow various referring instructions. 
This enhances the model's flexibility in understanding user intentions and instructions in different task contexts.
\citet{wan2024contrastive} propose to improve the grounding of vision-language models by contrastive region guidance. 
By guiding the model’s attention to relevant regions, MLLM can more accurately associate visual regions with corresponding textual instructions.
\citet{cho2024language} extend vision-language models to understand 3D environments, by improving spatial awareness and the understanding of object interactions in three-dimensional spaces.

\subsection{Visual Prompt Toolchain} 
To enable more complex multimodal understanding by multi-step or interactive reasoning, several methods aggregate various visual prompting methods as toolchains \cite{wu2024dettoolchain} to be called by the MLLM and assist individual reasoning sub-tasks.
\citet{zhou2024image} propose an image-of-thought method that can automatically determine each reasoning step's visual information extraction method and implement it as visual prompts, which prompt MLLM to follow a certain reasoning path and enable step-by-step multimodal reasoning.
\citet{tziafas2024towards} focus on adapting vision-language models for open-world grasping tasks by incorporating a list of visual prompting methods including open-end segmentation and object grounding to enable open-world grasping tasks.
To enable more transferable and generalizable visual prompts, 
\citet{sheng2024towards} create a more unified in-context learning method by integrating various contextual visual prompts into a unified representational space.
MineDreamer \cite{zhou2024minedreamer} further develops a versatile visual prompt generation method for imaginary visual scenes, which are consistent with current decision-making intention and visually express the next-step goal.

\subsection{Learnable and Soft Visual Prompt}
Learnable or soft visual prompts are employed to adapt the visual encoder in MLLMs, 
enabling more controlled and versatile use of visual prompts that are aligned with downstream tasks.
Such techniques are used in multimodal instruction tuning with visual instructions.
\citet{rezaei2024learning} investigates how visual prompts can be learned to guide the attention mechanisms in ViT.
\citet{li2023fine} fine-tune MLLMs to follow zero-shot demonstrative instructions using learnable visual prompts.
\citet{chen2023understanding} focus on better mapping visual inputs to corresponding labels through learned prompts.
For some specific and domain-oriented problems, 
\cite{ren2024you} develop a learnable visual prompting method as image watermarking to identify the image's copyright and ownership. 

At the same time, learnable visual prompts can also be transferable across MLLMs and downstream tasks.
VPGTrans \cite{zhang2024vpgtrans} proposes a transferable visual prompt generator, which adapts the pre-trained source MLLM to target MLLM with low cost in training data points and computation.
Memory-space visual prompt \cite{jie2024memory} injects learnable prompts in the key and value layers in the vision-transformer architecture, which enables efficient vision-language fine-tuning. 
\citet{wu2023few} also injects soft visual tokens as visual compositional operations, which are learned to better compose multimodal information with few-shot examples.
The black-box visual prompting method \cite{oh2023blackvip} focuses on robust transfer learning, where the visual prompts help models adapt to new tasks and domains without direct access to model parameters.

\section{Visual Perception} \label{sec:perception}

\subsection{Visual Grounding and Referring}
Recent visual prompting works have significantly improved MLLM's visual grounding and referring abilities.
Some works emphasize the importance of iterative feedback and multimodal interaction to refine semantic grounding, 
while others explore object-centric perception and the comprehension of visual relations.
To improve MLLM's regional grounding and object detection abilities,
SoM-LLaVA \cite{yan2024list} employs the Set-of-Mark (SoM) model to tag all the objects in the image and ask the model to list all the items.
InstructDET \cite{dang2023instructdet} and VTPrompt \cite{jiang2024joint} further enable multimodal grounding, 
which extracts object entities from the text and these objects' regional bounding boxes.

With a fine-grained visual grounding encoder, several works further use visual cues to guide MLLM's attention to relevant regions within images and achieve better regional referring abilities.
CRG \cite{wan2024contrastive} uses contrastive regional guidance to direct the model’s attention to specific areas of interest within an image, without model finetuning.
RelationVLM \cite{huang2024relationvlm} leverages visual prompts to enhance MLLM's understanding and reasoning about objects' spatial relations.
Shikra \cite{chen2023shikra} further applies to visual dialogue systems, where MLLM responds to referential cues within a dialogue, enabling more precise and context-aware interactions
In addition, several works aim to provide a comprehensive framework that incorporates various visual prompting methods in different granularity levels,
to enable more fine-grained and flexible multimodal interactions, including free-form visual prompt inputs \cite{lin2024draw} and feedback mechanisms \cite{liao2024can} on visual prompts.

\subsection{Multi-image and Video Understanding}
To improve the models' understanding of complex visual relationships and ensure that they can accurately reference and describe objects across diverse multi-image inputs,
several works propose visual prompts in multi-image inputs and novel evaluation benchmarks to test their effectiveness.
\citet{fan2024muffin} present a novel benchmark dataset with multipanel images to test MLLM's abilities in distinguishing objects across panels and navigating between different visual elements.
\citet{pan2024auto} leverage morph-token auto-encoding to enhance the model's capacity to process visual grounding across multiple images.
\citet{li2023fine} fine-tune MLLMs to follow in-context demonstrative instructions across multiple images.
In addition, AIM \cite{gao2024aim} proposes to dynamically adapt its grounding and referring abilities to accommodate new visual contexts across several images.

Several methods are also developed to allow MLLMs to identify specific regions of interest, improving their ability to handle complex and dynamic video content.
OmAgent \cite{zhang2024omagent} develops a visual prompting method to enable task division in video understanding, by annotating a series of visual features.
RACCooN \cite{yoon2024raccoon} uses visual prompts to guide MLLMs in identifying the target regions in the video for manipulation.
\citet{wu2024general} ground objects across videos, enabling the model to comprehend and refer to objects in dynamic scenes.

\subsection{3D Visual Understanding}
Recent works use visual prompting for better 3D visual understanding. 
\citet{li20243dmit} constructs an extensive dataset comprising instruction-responses pairs for 3D scenes and introduces 3DMIT for efficient prompt tuning while eliminating the alignment stage between 3D scenes and languages. 
DOrA \cite{wu2024dora} proposes a novel 3D visual grounding framework with Order-Aware referring. 
This method leverages LLM to infer ordered object series that used to guide the progressive feature refinement process. 

\citet{cho2024language} constructs a large-scale dataset named LV3D and introduces a new MLLM Cube-LLM pre-trained on the proposed dataset.
\citet{zhang2024agent3d} proposes Agent3D-Zero, which introduces novel visual prompts by employing bird's-eye view images 
and selecting viewpoints to unleash the MLLM's ability to observe 3D scenes.
3DAP~\cite{liu20233daxiesprompts} develops a novel visual prompting method that creates a 3D coordinate system a
nd additional annotation to empower GPT-4V to complete 3D spatial tasks.

\section{Compositional Reasoning} \label{sec:reason}
This section discusses how visual prompting enhances compositional and multimodal learning in MLLMs, 
enabling improvements in tasks like visual planning, reasoning, and action generation. 
We examine how visual prompts facilitate complex step-by-step reasoning, decision-making, and control over visual generation models, 
expanding their capabilities across diverse tasks.
We also review several frontier applications (Appendix \ref{sec:future}), which can be under-explored and lack sufficient solutions.

\subsection{Visual Planning}
Recent works demonstrate that visual prompting improves visual planning tasks. 
\citet{zhou2024image} proposes an Image-of-Thought(IoT) prompting method that 
compels MLLMs to automatically design visual and textual steps and leverages external image processing tools to generate a multi-model rationale series, 
which is used to assist MLLMs with complex visual reasoning tasks through a step-by-step process. 
OWG~\cite{tziafas2024towards} combines segmentation and grasp synthesis models, which unlocks the grounded world understanding through segmentation, grasp planning, and ranking. 
\citet{zhou2024minedreamer} introduces the Chain-of-Imagination (CoI) method and creates an embodied agent in Minecraft named MineDreamer. 
This method envisions the step-by-step process of executing instructions with the help of an LLM-enhanced diffusion model that translates imaginations into precise visual prompts to support the accurate generation of the agent's actions. 
BEVInstructor \cite{fan2024navigation} incorporates Bird's Eye View representations as visual prompts into MLLMs for navigation instruction generation. 
AO-Planner~\cite{chen2024affordances} achieves affordances-oriented motion planning and action decision-making with a VAP approach and a high-level PathAgent.

\subsection{Chain-of-thought}
To enable more complex image reasoning, recent works incorporate visual prompting with Chain-of-Thought methods. \citet{luan2024textcot} proposes a novel Chain-of-Thought framework for text-rich image understanding, named TextCoT. This method consists three stages including image overview for global information, coarse localization for estimating the section that encompasses the answer and fine-grained observation for furnishing precise answers. \citet{wu2024dettoolchain} proposes DetToolChain to unlock the potential of MLLMs in object detection task. This method involves using a "detection prompting toolkit," which includes visual processing and detection reasoning prompts, combined with a multimodal detection Chain-of-Thought method to reason the sequential implementation of the detection prompts.

\section{Model Training} \label{sec:train}
This section presents key approaches to align multimodal large language models (MLLMs) using visual prompting techniques,
including pre-training, fine-tuning, and instruction tuning, which aim to unify multi-modal prompts and improve cross-task transferability.
In addition to model training techniques, we also summarize evaluation datasets (Appendix \ref{sec:eval}), 
which inspire future work to develop more powerful visual prompting methods. 

\subsection{Pre-training}
To improve MLLM's ability on more fine-grained vision perception or reasoning tasks, a line of works focuses on designing better pre-training objectives including visual prompts. 
PSALM \cite{zhang2024psalm} extends the capabilities of MLLM to address various image segmentation tasks by incorporating a mask decoder and a flexible input schema. 
This approach unifies multiple segmentation tasks within a single model, supporting generic, referring, interactive, and open-vocabulary segmentation, while demonstrating strong performance on both in-domain and out-of-domain pixel-level segmentation tasks. 
OMG-LLaVA \cite{zhang2024omg} proposes a unified framework that bridges image-level, object-level, and pixel-level reasoning and understanding in a single model that combines a universal segmentation method as the visual encoder with an LLM, enabling flexible user interaction through various visual and text prompts. 
VisionLLM v2 \cite{wu2024visionllm} introduces an end-to-end generalist MLLM that unifies visual perception, understanding, and generation within a single framework. The model employs a novel "super link" technique to connect the central LLM with task-specific decoders, enabling flexible information transmission and end-to-end optimization across hundreds of vision and vision-language tasks.
UrbanVLP \cite{hao2024urbanvlp} proposes a vision-language pretraining framework for urban region profiling that integrates multi-granularity information from both satellite (macro-level) and street-view (micro-level) imagery, overcoming previous limitations. This method also incorporates an automatic text generation and calibration mechanism to produce high-quality textual descriptions of urban areas, enhancing interpretability.

\subsection{Fine-tuning}

\citet{zhang2024exploring} propose Transferable Visual Prompting (TVP), a method to improve the transferability of soft visual prompts which are a small amount of learnable parameters across different MLLMs for downstream tasks. 
\citet{lin2024rethinking} integrate fine-grained external knowledge such as OCR and segmentation into multimodal MLLMs through visual prompts, 
which embed fine-grained knowledge information directly into a spatial embedding map. 
CoLLaVO \cite{lee2024collavo} enhances MLLMs' object-level image understanding through a visual prompt called Crayon Prompt, 
which is derived from panoptic color maps generated by a panoptic segmentation model. 
CityLLaVA \cite{duan2024cityllava} introduces an efficient fine-tuning framework for MLLM designed for urban scenarios which incorporates visual prompt engineering techniques, including bounding box-guided, view selection, and global-local joint views. 
ViP-LLaVA~\cite{cai2024vip} is enabled to understand arbitrary visual prompts, which is trained by directly overlaying visual markers onto images.
ImageBrush \cite{yang2024imagebrush} introduces a framework for exemplar-based image manipulation that learns visual in-context instructions without language prompts. 

Explicit Visual Prompting (EVP) \cite{liu2023explicit} proposes a unified approach for low-level structure segmentation tasks with a frozen pre-trained vision transformer backbone and introduces task-specific soft prompts derived from frozen patch embeddings and high-frequency image components. 
BlackVIP \cite{oh2023blackvip} adapts large pre-trained models with a Coordinator to generate soft visual prompts and SPSA-GC for efficient gradient estimation, enabling robust few-shot adaptation across diverse domains. 
Iterative Label Mapping-based Visual Prompting (ILM-VP) \cite{chen2023understanding} improves the accuracy and interpretability of soft visual prompting by jointly optimizing input patterns and label mapping through bi-level optimization. 
MemVP \cite{jie2024memory} efficiently combines pre-trained vision encoders and language models for vision-language tasks by injecting visual information directly into the feed-forward network weights of MLLMs, treating them as additional factual knowledge. 
VPG-C \cite{li2023fine} enhances visual prompting in MLLMs by completing missing visual details to better comprehend demonstrative instructions with interleaved multimodal context. 
It extends traditional Visual Prompt Generators by using LLM-guided, context-aware visual feature extraction to create more comprehensive visual prompts. 

\subsection{Instruction Tuning}
Instruction tuning has proved to effectively improve the overall ability of both text-only LLMs and MLLMs such as instruction following and structured output \cite{ouyang2022training, wang2022self, liu2024visual}. 
For MLLMs with a focus on visual prompts, AnyRef \cite{he2024multi} introduces a unified referring representation that enables the MLLM to handle diverse input modalities and visual prompts (text, bounding boxes, images, audio) through instruction tuning. This model uses special tokens and prompts to format multi-modal inputs, allowing it to process various referring formats consistently. 
A refocusing mechanism enhances mask embeddings by incorporating grounded textual embeddings, improving segmentation accuracy. AnyRef combines vision and audio encoders with an LLM, using projection layers to align different modalities in the language space. The model is instruction-tuned end-to-end with a combination of text loss and mask loss, enabling it to generate both textual descriptions and pixel-level segmentation in response to multi-modal prompts.

\section{In-context and Few-shot Learning } \label{sec:icl}
Beyond methods that optimize performance using single data points as input, some works focus on enhancing in-context learning (ICL) with visual prompts.
Image-of-Thought (IoT) prompting \cite{zhou2024image} is a train-free approach to enhance MLLMs for visual question-answering tasks by integrating discrete image processing actions.
IoT enables MLLMs to automatically design and extract visual rationales step-by-step, combining them with textual rationales to improve both accuracy and interpretability. 
CRG \cite{wan2024contrastive} is a training-free method that improves visual grounding in MLLMs by contrasting model outputs with and without specific image regions masked which guides models to focus on relevant image areas. 
AIM \cite{gao2024aim} enables any MLLM to perform efficient ICL by aggregating image information from demonstrations into the latent space of corresponding textual labels which reduces memory costs by discarding visual tokens after aggregation, approximating multimodal ICL prompts to contain only a single query image. 
I2L\cite{Wang2024AllIA} combines demonstrations, visual cues, and reasoning into a single image to enhance multimodal models' performance on complex tasks through ICL. I2L-Hybrid extends this by automatically selecting between I2L and other in-context learning methods for each task instance.

Few-shot learning through visual prompts can also improve the capabilities of MLLMs with minimum computational cost and better data efficiency. 
CoMM \cite{chen2024comm} proposes a high-quality coherent interleaved image-text dataset designed to enhance the generation capabilities of MLLMs and investigate their in-context learning ability. 
M2oEGPT \cite{sheng2024towards} propose an ICL framework by using multimodal quantization and unified embedding to enable joint learning of multimodal data in a general token embedding space, combining an autoregressive transformer with a Mixture of Experts (MoEs) for stable multi-task co-training. 
Partial2Global \cite{xu2024towards} select optimal in-context examples in visual ICL through a transformer-based list-wise ranker to compare multiple alternative samples and a consistency-aware ranking aggregator to achieve globally consistent ranking. 
\citet{hossain2024visual} introduces learnable visual prompts for both base and novel classes on semantic segmentation, along with a novel-to-base causal attention mechanism that allows novel prompts to be contextualized by base prompts without degrading base class performance. 
Emu2 \cite{sun2024generative} is MLLM trained to predict the next element in diverse multimodal sequences. Its unified architecture enables strong multimodal in-context learning abilities, allowing it to quickly adapt to new tasks with just a few examples. 

\section{Evaluation } \label{sec:eval}

\begin{table*}[ht]
\centering
\small
\begin{tabular}{l|ccc|ccc|cc}
\toprule
\textbf{Reference} & \textbf{SP} & \textbf{TP} & \textbf{GP} & \textbf{Image}  & \textbf{Video} & \textbf{Audio} & \textbf{Manual} & \textbf{Automatic} \\
\midrule
MDVP-Bench \cite{lin2024draw} & \checkmark & \checkmark & \checkmark & \checkmark & & & \checkmark & \checkmark \\
A3VLM \cite{huang2024a3vlm} & \checkmark & & & \checkmark & & & & \checkmark \\
VLM Feedback \cite{li2023vrptest} & \checkmark & & & \checkmark & \checkmark & & \checkmark & \checkmark\\
GPT-4V Challenger \cite{fu2023challenger} & \checkmark & & \checkmark & \checkmark & & & \\
EarthMarker \cite{zhang2024earthmarker} & \checkmark & \checkmark & \checkmark & \checkmark & & & \checkmark \\
RACCooN \cite{yoon2024raccoon} & & & \checkmark & & \checkmark & & \checkmark & \checkmark \\
Safety of MLLMs\cite{liu2024safety} & & & \checkmark & \checkmark & \checkmark & \checkmark & \\
GLEE \cite{wu2024general} & \checkmark & \checkmark & & \checkmark & \checkmark & \\
AutoAD-Zero \cite{xie2024autoad} & \checkmark & & \checkmark & \checkmark & & \checkmark & & \checkmark \\
MultipanelVQA \cite{fan2024muffin} & & & \checkmark & \checkmark & & & \\
MM-Vid \cite{lin2023mm} & \checkmark & \checkmark & \checkmark & \checkmark & \checkmark & \checkmark & \checkmark \\
Groma \cite{ma2024groma} & \checkmark & \checkmark & & \checkmark & & & \checkmark & \checkmark \\
\bottomrule
\end{tabular}
\vspace{1em}
\captionsetup{width=\linewidth}
\caption{We compare different benchmarks and training datasets, and they are each grouped into three different criteria--Semantic Prompting (SP), Textual Prompting (TP), and Generative Prompting (GP). Then, based on the different modalities, they can be classified if they contain pixel-level images (Image), video encoding and decoding (Video), and if they are supplemented by an audio transcript (Audio). Finally, the last categorization determines if the specified method visual prompting is done manually (Manual), automated (Automatic), or a combination of both.}
\label{tab:comparison_new}
\end{table*}
This section explores and compares the current MLLM visual prompting training datasets and benchmarks, as visualized in Section 7.1. The three main categories for the visual prompting techniques are Semantic Prompting (SP), Textual Prompting (TP), and GP (Generative Prompting).

The datasets and benchmarks that fall into the Semantic Prompting (SP) utilize high-level descriptions to help the model understand the semantic relationships present in the data. Some examples incude creating bounding boxes \cite{huang2024a3vlm,wu2024general}, labeling regions of interest \cite{li2023vrptest}, and tagging objects \cite{lin2024draw,zhang2024earthmarker}. Another general method is Textual Prompting (TP) where either user or LLM generated text is supplemented into the model input that relates the visual aspects in the image. Image and video descriptions can be generated and used as a visual prompt \cite{lin2023mm}, drawing relationships and descriptions on the image itself in order to add location-specific analysis \cite{lin2024draw,wu2024general}, and embedding localization into image tokenization \cite{ma2024groma}. Given the extensive effort required for manual visual prompting in MLLMs, some techniques have adopted automatic generation methods to streamline the visual prompting process using Generative Prompting (GP). Automatic modality conversion uses LLMs to generate  text from images/videos and vice versa for users to easily modify and cater the prompts \cite{yoon2024raccoon}. Audio descriptions are generated and then summarized by an LLM [\cite{xie2024autoad}, [\cite{lin2023mm}]]. Similarly, generation is used to create difficult and unique benchmarks to assess the capabilities and weaknesses of specific models [\cite{fan2024muffin}].

The final taxonomy system distinguishes between those visual prompting techniques that are done manually between those that are done automatically. The manual techniques offer precision and customization, but in turn sacrifice time and efficiency. They are suitable for tasks that are smaller scale and require detail. Automatic techniques provide scalability and productivity--they work well with large scale tasks that do not require fine-grained accuracy. Some techniques apply a combination of these \cite{lin2024draw,li2023vrptest,yoon2024raccoon,ma2024groma} and those that do not have either checked were either training datasets or surveys themselves \cite{liu2024safety,fan2024muffin,fu2023challenger}.

\section{Frontier Applications } \label{sec:future}
\subsection{Jailbreaking \& Safety}
While visual prompting enables fine-grained instructions to MLLMs for better response generation, it can also be intentionally designed to expose critical safety issues of MLLMs \cite{liu2024safety, ni2024responsible}.
Several works have explored jailbreaking of MLLMs with visual prompts. 
Instead of feeding harmful textual instructions directly, \citet{gong2023figstep} converts them into images through typography and feeds them to MLLMs as visual prompts.
The results show that even if the underlying LLM has been aligned for safety, visual prompting opens a new jailbreaking surface generating harmful responses.

To further expose the safety problems of MLLMs for red-teaming, multimodal jailbreaking prompts combining both textual and visual instructions are also studied.
\citet{ying2024jailbreak} first embed harmful perturbation in the visual prompt and then optimize the textual prompt through LLM reasoning on the harmful intent in the image. 
Meanwhile, \citet{liu2024arondight} utilize a red-team MLLM and a red-team LLM guided by reinforcement learning to automatically generate visual and textual jailbrearking prompts respectively. 
Their results suggest that multimodal prompts could lead to stronger attack on MLLMs that fuse multimodal input features.
Furthermore, \citet{gu2024agent} observe a more severe safety issue of infectious jailbreark in multi-agent MLLM environments. With an adversarial image simply jailbreaking one agent and without any further intervention, almost all agents will start exhibit harmful behaviors in an exponential infection rate during multi-agent interaction.

\subsection{Hallucination}
The more fine-grained visual contexts provided with visual prompting are also useful for multimodal hallucination mitigation.
To address the issue that MLLMs' textual outputs are often not grounded in the reference images, \citet{favero2024multi} propose a mutual-information decoding strategy to amplify the influence of visual prompts on model generation.
To reduce MLLMs' object hallucination and enhance fine-grained understanding in object-oriented perception tasks, \citet{jiang2024joint} develop a prompting strategy jointly utilizing visual and textual prompts.
A specialized detection model is employed to highlight relevant visual objects and visual prompts based on the key concepts extracted from textual prompts.
While previous works mostly focus on single-object hallucination, \citet{chen2024multi} utilize visual referring prompts to evaluate multi-object hallucination of MLLMs.
The results show MLLMs tend to experience more hallucinations when tasked with focusing multiple objects at the same time and authors suggest probing objects individually in visual prompts to enhance performances.

\subsection{Debiasing}
Despite the impressive capabilities of MLLMs, the biases and robustness of them remain a crucial challenge where models tend to utilize spurious correlations between input and target variables for predictions leading to potential social biases on certain topics, e.g., gender and racial biases \cite{ye2024mm}.
As visual prompting enables more fine-grained understanding of visual objects and relationships, it serves as a promising solution to mitigate potential biases in MLLMs' generations by grounding the outputs with essential visual information and thus avoiding spurious correlations of non-essential inputs.
It may also enhance the causal understanding of MLLMs between objects from the same modality and across different modalities for generating more robust and grounded responses.

\subsection{Visual Generation}
Visual generation models, especially text-to-image diffusion models \cite{rombach2022high}, are becoming popular. Considering large-scale pre-trained diffusion models as MLLMs broadly, visual prompting plays an important role in controlling the generation and enable diffusion models for unseen visual tasks. \citet{zhang2023adding}, \citet{mou2024t2i} propose ControlNet and T2I Adapter, which take various visual prompts for spatial control in image generation. In this survey, we discuss works that focus on visual prompting instead of controllable generation \cite{cao2024controllable} in general. Prompt Diffusion \cite{wang2023context} proposes a diffusion-based generative model that takes a novel vision-language prompts and outputs the target images, which unlocks the ability of in-context generation after fine-tuned on six visual tasks.
ImageBrush \cite{yang2024imagebrush} proposes to achieve adaptive image manipulation under the instruction of a pair of exemplar demonstrations in order to address the issue of language ambiguity in image editing task. 
MPerceiver \cite{ai2024multimodal} introduces a multi-modal prompt learning approach using generative priors of diffusion models to enhance the all-in-one image restoration. 
\citet{chen2024vp3d} proposes VP3D, which leverages rich knowledge in 2D visual prompts to improve text-to-3D generation quality and trigger a new task of stylized text-to-3D generation. 
PromptCharm \cite{wang2024promptcharm} proposes an interaction system that supports text-to-image creation through multi-modal prompting and image refinement, which suggests the necessity of visual prompting for better image creation.

\section{Conclusion }
In this survey, we provide the first comprehensive review of visual prompting methods in MLLMs. 
We categorized various visual prompting techniques and discussed their generation processes,
examining their integration into MLLMs for enhanced visual reasoning and perception.
Our work also examines existing training and in-context learning methods in MLLMs with visual prompting. 
We inspire future directions that leverage visual prompts for better MLLM compositional reasoning.

\section{Limitations }
While our survey offers a comprehensive overview, it may be limited by the rapidly evolving nature of the field and potential gaps in the available literature. 
Future work should focus on expanding the scope of visual prompts and refining alignment techniques to further enhance MLLM capabilities.

\bibliography{ref}

\begin{thebibliography}{108}
\providecommand{\natexlab}[1]{#1}

\bibitem[{Ai et~al.(2024)Ai, Huang, Zhou, Wang, and He}]{ai2024multimodal}
Yuang Ai, Huaibo Huang, Xiaoqiang Zhou, Jiexiang Wang, and Ran He. 2024.
\newblock Multimodal prompt perceiver: Empower adaptiveness generalizability and fidelity for all-in-one image restoration.
\newblock In \emph{Proceedings of the IEEE/CVF Conference on Computer Vision and Pattern Recognition}, pages 25432--25444.

\bibitem[{Bai et~al.(2024)Bai, Wang, Xiao, He, Han, Zhang, and Shou}]{bai2024hallucination}
Zechen Bai, Pichao Wang, Tianjun Xiao, Tong He, Zongbo Han, Zheng Zhang, and Mike~Zheng Shou. 2024.
\newblock Hallucination of multimodal large language models: A survey.
\newblock \emph{arXiv preprint arXiv:2404.18930}.

\bibitem[{Bar et~al.(2022)Bar, Gandelsman, Darrell, Globerson, and Efros}]{bar2022visual}
Amir Bar, Yossi Gandelsman, Trevor Darrell, Amir Globerson, and Alexei Efros. 2022.
\newblock Visual prompting via image inpainting.
\newblock \emph{Advances in Neural Information Processing Systems}, 35:25005--25017.

\bibitem[{Brazil et~al.(2023)Brazil, Kumar, Straub, Ravi, Johnson, and Gkioxari}]{brazil2023omni3d}
Garrick Brazil, Abhinav Kumar, Julian Straub, Nikhila Ravi, Justin Johnson, and Georgia Gkioxari. 2023.
\newblock Omni3d: A large benchmark and model for 3d object detection in the wild.
\newblock In \emph{Proceedings of the IEEE/CVF conference on computer vision and pattern recognition}, pages 13154--13164.

\bibitem[{Cai et~al.(2024)Cai, Liu, Mustikovela, Meyer, Chai, Park, and Lee}]{cai2024vip}
Mu~Cai, Haotian Liu, Siva~Karthik Mustikovela, Gregory~P Meyer, Yuning Chai, Dennis Park, and Yong~Jae Lee. 2024.
\newblock Vip-llava: Making large multimodal models understand arbitrary visual prompts.
\newblock In \emph{Proceedings of the IEEE/CVF Conference on Computer Vision and Pattern Recognition}, pages 12914--12923.

\bibitem[{Cao et~al.(2024)Cao, Zhou, Song, and Yang}]{cao2024controllable}
Pu~Cao, Feng Zhou, Qing Song, and Lu~Yang. 2024.
\newblock Controllable generation with text-to-image diffusion models: A survey.
\newblock \emph{arXiv preprint arXiv:2403.04279}.

\bibitem[{Chen et~al.(2023{\natexlab{a}})Chen, Yao, Chen, Zhang, and Liu}]{chen2023understanding}
Aochuan Chen, Yuguang Yao, Pin-Yu Chen, Yihua Zhang, and Sijia Liu. 2023{\natexlab{a}}.
\newblock Understanding and improving visual prompting: A label-mapping perspective.
\newblock In \emph{Proceedings of the IEEE/CVF Conference on Computer Vision and Pattern Recognition}, pages 19133--19143.

\bibitem[{Chen et~al.(2024{\natexlab{a}})Chen, Lin, Liu, Liang, and Wong}]{chen2024affordances}
Jiaqi Chen, Bingqian Lin, Xinmin Liu, Xiaodan Liang, and Kwan-Yee~K Wong. 2024{\natexlab{a}}.
\newblock Affordances-oriented planning using foundation models for continuous vision-language navigation.
\newblock \emph{arXiv preprint arXiv:2407.05890}.

\bibitem[{Chen et~al.(2023{\natexlab{b}})Chen, Zhang, Zeng, Zhang, Zhu, and Zhao}]{chen2023shikra}
Keqin Chen, Zhao Zhang, Weili Zeng, Richong Zhang, Feng Zhu, and Rui Zhao. 2023{\natexlab{b}}.
\newblock Shikra: Unleashing multimodal llm's referential dialogue magic.
\newblock \emph{arXiv preprint arXiv:2306.15195}.

\bibitem[{Chen et~al.(2024{\natexlab{b}})Chen, Li, Yang, Wen, Yang, Gao, Wu, and Chen}]{chen2024comm}
Wei Chen, Lin Li, Yongqi Yang, Bin Wen, Fan Yang, Tingting Gao, Yu~Wu, and Long Chen. 2024{\natexlab{b}}.
\newblock Comm: A coherent interleaved image-text dataset for multimodal understanding and generation.
\newblock \emph{arXiv preprint arXiv:2406.10462}.

\bibitem[{Chen et~al.(2024{\natexlab{c}})Chen, Ma, Zhang, Xu, Qian, Yang, Fouhey, and Chai}]{chen2024multi}
Xuweiyi Chen, Ziqiao Ma, Xuejun Zhang, Sihan Xu, Shengyi Qian, Jianing Yang, David~F Fouhey, and Joyce Chai. 2024{\natexlab{c}}.
\newblock Multi-object hallucination in vision-language models.
\newblock \emph{arXiv preprint arXiv:2407.06192}.

\bibitem[{Chen et~al.(2024{\natexlab{d}})Chen, Pan, Yang, Yao, and Mei}]{chen2024vp3d}
Yang Chen, Yingwei Pan, Haibo Yang, Ting Yao, and Tao Mei. 2024{\natexlab{d}}.
\newblock Vp3d: Unleashing 2d visual prompt for text-to-3d generation.
\newblock In \emph{Proceedings of the IEEE/CVF Conference on Computer Vision and Pattern Recognition}, pages 4896--4905.

\bibitem[{Cho et~al.(2024)Cho, Ivanovic, Cao, Schmerling, Wang, Weng, Li, You, Kr{\"a}henb{\"u}hl, Wang et~al.}]{cho2024language}
Jang~Hyun Cho, Boris Ivanovic, Yulong Cao, Edward Schmerling, Yue Wang, Xinshuo Weng, Boyi Li, Yurong You, Philipp Kr{\"a}henb{\"u}hl, Yan Wang, et~al. 2024.
\newblock Language-image models with 3d understanding.
\newblock \emph{arXiv preprint arXiv:2405.03685}.

\bibitem[{Dang et~al.(2023)Dang, Feng, Zhang, Ge, Song, Gong, Liu, Chen, Zhu, Zhao et~al.}]{dang2023instructdet}
Ronghao Dang, Jiangyan Feng, Haodong Zhang, Chongjian Ge, Lin Song, Lijun Gong, Chengju Liu, Qijun Chen, Feng Zhu, Rui Zhao, et~al. 2023.
\newblock Instructdet: Diversifying referring object detection with generalized instructions.
\newblock \emph{arXiv preprint arXiv:2310.05136}.

\bibitem[{Duan et~al.(2024)Duan, Cheng, Xu, Wu, Zhang, Ye, and Xie}]{duan2024cityllava}
Zhizhao Duan, Hao Cheng, Duo Xu, Xi~Wu, Xiangxie Zhang, Xi~Ye, and Zhen Xie. 2024.
\newblock Cityllava: Efficient fine-tuning for vlms in city scenario.
\newblock In \emph{Proceedings of the IEEE/CVF Conference on Computer Vision and Pattern Recognition}, pages 7180--7189.

\bibitem[{Fan et~al.(2024{\natexlab{a}})Fan, Liu, Wang, and Yang}]{fan2024navigation}
Sheng Fan, Rui Liu, Wenguan Wang, and Yi~Yang. 2024{\natexlab{a}}.
\newblock Navigation instruction generation with bev perception and large language models.
\newblock \emph{arXiv preprint arXiv:2407.15087}.

\bibitem[{Fan et~al.(2024{\natexlab{b}})Fan, Ding, Kuo, Jiang, Zhao, Guan, Yang, Zhang, and Wang}]{fan2024read}
Yue Fan, Lei Ding, Ching-Chen Kuo, Shan Jiang, Yang Zhao, Xinze Guan, Jie Yang, Yi~Zhang, and Xin~Eric Wang. 2024{\natexlab{b}}.
\newblock Read anywhere pointed: Layout-aware gui screen reading with tree-of-lens grounding.
\newblock \emph{arXiv preprint arXiv:2406.19263}.

\bibitem[{Fan et~al.(2024{\natexlab{c}})Fan, Gu, Zhou, Yan, Jiang, Kuo, Guan, and Wang}]{fan2024muffin}
Yue Fan, Jing Gu, Kaiwen Zhou, Qianqi Yan, Shan Jiang, Ching-Chen Kuo, Xinze Guan, and Xin~Eric Wang. 2024{\natexlab{c}}.
\newblock Muffin or chihuahua? challenging large vision-language models with multipanel vqa.
\newblock \emph{arXiv preprint arXiv:2401.15847}.

\bibitem[{Favero et~al.(2024)Favero, Zancato, Trager, Choudhary, Perera, Achille, Swaminathan, and Soatto}]{favero2024multi}
Alessandro Favero, Luca Zancato, Matthew Trager, Siddharth Choudhary, Pramuditha Perera, Alessandro Achille, Ashwin Swaminathan, and Stefano Soatto. 2024.
\newblock Multi-modal hallucination control by visual information grounding.
\newblock In \emph{Proceedings of the IEEE/CVF Conference on Computer Vision and Pattern Recognition}, pages 14303--14312.

\bibitem[{Fu et~al.(2023)Fu, Zhang, Lin, Wang, Gao, Luo, Huang, Zhang, Qiu, Ye et~al.}]{fu2023challenger}
Chaoyou Fu, Renrui Zhang, Haojia Lin, Zihan Wang, Timin Gao, Yongdong Luo, Yubo Huang, Zhengye Zhang, Longtian Qiu, Gaoxiang Ye, et~al. 2023.
\newblock A challenger to gpt-4v? early explorations of gemini in visual expertise.
\newblock \emph{arXiv preprint arXiv:2312.12436}.

\bibitem[{Gao et~al.(2024)Gao, Qiao, Cao, Wang, and Li}]{gao2024aim}
Jun Gao, Qian Qiao, Ziqiang Cao, Zili Wang, and Wenjie Li. 2024.
\newblock Aim: Let any multi-modal large language models embrace efficient in-context learning.
\newblock \emph{arXiv preprint arXiv:2406.07588}.

\bibitem[{Girshick(2015)}]{girshick2015fast}
Ross Girshick. 2015.
\newblock Fast r-cnn.
\newblock In \emph{Proceedings of the IEEE international conference on computer vision}, pages 1440--1448.

\bibitem[{Gong et~al.(2023)Gong, Ran, Liu, Wang, Cong, Wang, Duan, and Wang}]{gong2023figstep}
Yichen Gong, Delong Ran, Jinyuan Liu, Conglei Wang, Tianshuo Cong, Anyu Wang, Sisi Duan, and Xiaoyun Wang. 2023.
\newblock Figstep: Jailbreaking large vision-language models via typographic visual prompts.
\newblock \emph{arXiv preprint arXiv:2311.05608}.

\bibitem[{Gu et~al.(2023)Gu, Han, Chen, Beirami, He, Zhang, Liao, Qin, Tresp, and Torr}]{gu2023systematic}
Jindong Gu, Zhen Han, Shuo Chen, Ahmad Beirami, Bailan He, Gengyuan Zhang, Ruotong Liao, Yao Qin, Volker Tresp, and Philip Torr. 2023.
\newblock A systematic survey of prompt engineering on vision-language foundation models.
\newblock \emph{arXiv preprint arXiv:2307.12980}.

\bibitem[{Gu et~al.(2024)Gu, Zheng, Pang, Du, Liu, Wang, Jiang, and Lin}]{gu2024agent}
Xiangming Gu, Xiaosen Zheng, Tianyu Pang, Chao Du, Qian Liu, Ye~Wang, Jing Jiang, and Min Lin. 2024.
\newblock Agent smith: A single image can jailbreak one million multimodal llm agents exponentially fast.
\newblock \emph{arXiv preprint arXiv:2402.08567}.

\bibitem[{Hao et~al.(2024)Hao, Chen, Yan, Zhong, Wang, Wen, and Liang}]{hao2024urbanvlp}
Xixuan Hao, Wei Chen, Yibo Yan, Siru Zhong, Kun Wang, Qingsong Wen, and Yuxuan Liang. 2024.
\newblock Urbanvlp: A multi-granularity vision-language pre-trained foundation model for urban indicator prediction.
\newblock \emph{arXiv preprint arXiv:2403.16831}.

\bibitem[{He et~al.(2024)He, Wang, Wang, Lu, He, Lan, Luo, and Xie}]{he2024multi}
Junwen He, Yifan Wang, Lijun Wang, Huchuan Lu, Jun-Yan He, Jin-Peng Lan, Bin Luo, and Xuansong Xie. 2024.
\newblock Multi-modal instruction tuned llms with fine-grained visual perception.
\newblock In \emph{Proceedings of the IEEE/CVF Conference on Computer Vision and Pattern Recognition}, pages 13980--13990.

\bibitem[{Hossain et~al.(2024)Hossain, Siam, Sigal, and Little}]{hossain2024visual}
Mir Rayat~Imtiaz Hossain, Mennatullah Siam, Leonid Sigal, and James~J Little. 2024.
\newblock Visual prompting for generalized few-shot segmentation: A multi-scale approach.
\newblock In \emph{Proceedings of the IEEE/CVF Conference on Computer Vision and Pattern Recognition}, pages 23470--23480.

\bibitem[{Huang et~al.(2024{\natexlab{a}})Huang, Chang, Liu, Zhu, Dong, Gao, Boularias, and Li}]{huang2024a3vlm}
Siyuan Huang, Haonan Chang, Yuhan Liu, Yimeng Zhu, Hao Dong, Peng Gao, Abdeslam Boularias, and Hongsheng Li. 2024{\natexlab{a}}.
\newblock A3vlm: Actionable articulation-aware vision language model.
\newblock \emph{arXiv preprint arXiv:2406.07549}.

\bibitem[{Huang et~al.(2024{\natexlab{b}})Huang, Liu, Guo, and Gong}]{huang2024visual}
Wen Huang, Hongbin Liu, Minxin Guo, and Neil~Zhenqiang Gong. 2024{\natexlab{b}}.
\newblock Visual hallucinations of multi-modal large language models.
\newblock \emph{arXiv preprint arXiv:2402.14683}.

\bibitem[{Huang et~al.(2024{\natexlab{c}})Huang, Zhang, Zha, Lu, and Guo}]{huang2024relationvlm}
Zhipeng Huang, Zhizheng Zhang, Zheng-Jun Zha, Yan Lu, and Baining Guo. 2024{\natexlab{c}}.
\newblock Relationvlm: Making large vision-language models understand visual relations.
\newblock \emph{arXiv preprint arXiv:2403.12801}.

\bibitem[{Jia et~al.(2024)Jia, Zhang, Yu, Jiao, and Jiang}]{jia2024describe}
Mengzhao Jia, Zhihan Zhang, Wenhao Yu, Fangkai Jiao, and Meng Jiang. 2024.
\newblock Describe-then-reason: Improving multimodal mathematical reasoning through visual comprehension training.
\newblock \emph{arXiv preprint arXiv:2404.14604}.

\bibitem[{Jiang et~al.(2024)Jiang, Zhang, Zhou, Jin, Feng, Wu, and Liu}]{jiang2024joint}
Songtao Jiang, Yan Zhang, Chenyi Zhou, Yeying Jin, Yang Feng, Jian Wu, and Zuozhu Liu. 2024.
\newblock Joint visual and text prompting for improved object-centric perception with multimodal large language models.
\newblock \emph{arXiv preprint arXiv:2404.04514}.

\bibitem[{Jie et~al.(2024)Jie, Tang, Ding, Deng, Han, and Wang}]{jie2024memory}
Shibo Jie, Yehui Tang, Ning Ding, Zhi-Hong Deng, Kai Han, and Yunhe Wang. 2024.
\newblock Memory-space visual prompting for efficient vision-language fine-tuning.
\newblock \emph{arXiv preprint arXiv:2405.05615}.

\bibitem[{Kirillov et~al.(2023)Kirillov, Mintun, Ravi, Mao, Rolland, Gustafson, Xiao, Whitehead, Berg, Lo et~al.}]{kirillov2023segment}
Alexander Kirillov, Eric Mintun, Nikhila Ravi, Hanzi Mao, Chloe Rolland, Laura Gustafson, Tete Xiao, Spencer Whitehead, Alexander~C Berg, Wan-Yen Lo, et~al. 2023.
\newblock Segment anything.
\newblock In \emph{Proceedings of the IEEE/CVF International Conference on Computer Vision}, pages 4015--4026.

\bibitem[{Lee et~al.(2024)Lee, Park, Kim, and Ro}]{lee2024collavo}
Byung-Kwan Lee, Beomchan Park, Chae~Won Kim, and Yong~Man Ro. 2024.
\newblock Collavo: Crayon large language and vision model.
\newblock \emph{arXiv preprint arXiv:2402.11248}.

\bibitem[{Lei et~al.(2024{\natexlab{a}})Lei, Yang, Chen, Li, and Liu}]{lei2024scaffolding}
Xuanyu Lei, Zonghan Yang, Xinrui Chen, Peng Li, and Yang Liu. 2024{\natexlab{a}}.
\newblock Scaffolding coordinates to promote vision-language coordination in large multi-modal models.
\newblock \emph{arXiv preprint arXiv:2402.12058}.

\bibitem[{Lei et~al.(2024{\natexlab{b}})Lei, Li, Li, Cao, and Shan}]{lei2024prompt}
Yiming Lei, Jingqi Li, Zilong Li, Yuan Cao, and Hongming Shan. 2024{\natexlab{b}}.
\newblock Prompt learning in computer vision: a survey.
\newblock \emph{Frontiers of Information Technology \& Electronic Engineering}, 25(1):42--63.

\bibitem[{Li et~al.(2023{\natexlab{a}})Li, Pan, Ge, Gao, Ji, Zhang, Chua, Tang, Zhang, and Zhuang}]{li2023fine}
Juncheng Li, Kaihang Pan, Zhiqi Ge, Minghe Gao, Wei Ji, Wenqiao Zhang, Tat-Seng Chua, Siliang Tang, Hanwang Zhang, and Yueting Zhuang. 2023{\natexlab{a}}.
\newblock Fine-tuning multimodal llms to follow zero-shot demonstrative instructions.
\newblock In \emph{The Twelfth International Conference on Learning Representations}.

\bibitem[{Li et~al.(2023{\natexlab{b}})Li, Li, Savarese, and Hoi}]{li2023blip}
Junnan Li, Dongxu Li, Silvio Savarese, and Steven Hoi. 2023{\natexlab{b}}.
\newblock Blip-2: Bootstrapping language-image pre-training with frozen image encoders and large language models.
\newblock In \emph{International conference on machine learning}, pages 19730--19742. PMLR.

\bibitem[{Li(2023)}]{li2023practical}
Yinheng Li. 2023.
\newblock A practical survey on zero-shot prompt design for in-context learning.
\newblock \emph{arXiv preprint arXiv:2309.13205}.

\bibitem[{Li et~al.(2024)Li, Zhang, Wang, Ren, Xu, Ma, and Liu}]{li20243dmit}
Zeju Li, Chao Zhang, Xiaoyan Wang, Ruilong Ren, Yifan Xu, Ruifei Ma, and Xiangde Liu. 2024.
\newblock 3dmit: 3d multi-modal instruction tuning for scene understanding.
\newblock \emph{arXiv preprint arXiv:2401.03201}.

\bibitem[{Li et~al.(2023{\natexlab{c}})Li, Wang, Liu, Ma, Wu, Wang, and Gao}]{li2023vrptest}
Zongjie Li, Chaozheng Wang, Chaowei Liu, Pingchuan Ma, Daoyuan Wu, Shuai Wang, and Cuiyun Gao. 2023{\natexlab{c}}.
\newblock Vrptest: Evaluating visual referring prompting in large multimodal models.
\newblock \emph{arXiv preprint arXiv:2312.04087}.

\bibitem[{Liao et~al.(2024)Liao, Mahmood, Fidler, and Acuna}]{liao2024can}
Yuan-Hong Liao, Rafid Mahmood, Sanja Fidler, and David Acuna. 2024.
\newblock Can feedback enhance semantic grounding in large vision-language models?
\newblock \emph{arXiv preprint arXiv:2404.06510}.

\bibitem[{Lin et~al.(2023)Lin, Ahmed, Li, Lin, Azarnasab, Yang, Wang, Liang, Liu, Lu et~al.}]{lin2023mm}
Kevin Lin, Faisal Ahmed, Linjie Li, Chung-Ching Lin, Ehsan Azarnasab, Zhengyuan Yang, Jianfeng Wang, Lin Liang, Zicheng Liu, Yumao Lu, et~al. 2023.
\newblock Mm-vid: Advancing video understanding with gpt-4v (ision).
\newblock \emph{arXiv preprint arXiv:2310.19773}.

\bibitem[{Lin et~al.(2024{\natexlab{a}})Lin, Wei, An, Gao, Zou, Luo, Huang, Zhang, and Li}]{lin2024draw}
Weifeng Lin, Xinyu Wei, Ruichuan An, Peng Gao, Bocheng Zou, Yulin Luo, Siyuan Huang, Shanghang Zhang, and Hongsheng Li. 2024{\natexlab{a}}.
\newblock Draw-and-understand: Leveraging visual prompts to enable mllms to comprehend what you want.
\newblock \emph{arXiv preprint arXiv:2403.20271}.

\bibitem[{Lin et~al.(2024{\natexlab{b}})Lin, Li, Chen, Xu, Clark, Torr, and Yuan}]{lin2024rethinking}
Yuanze Lin, Yunsheng Li, Dongdong Chen, Weijian Xu, Ronald Clark, Philip Torr, and Lu~Yuan. 2024{\natexlab{b}}.
\newblock Rethinking visual prompting for multimodal large language models with external knowledge.
\newblock \emph{arXiv preprint arXiv:2407.04681}.

\bibitem[{Liu et~al.(2023{\natexlab{a}})Liu, Dong, Zhang, Luo, Gao, Huang, Gong, and Wang}]{liu20233daxiesprompts}
Dingning Liu, Xiaomeng Dong, Renrui Zhang, Xu~Luo, Peng Gao, Xiaoshui Huang, Yongshun Gong, and Zhihui Wang. 2023{\natexlab{a}}.
\newblock 3daxiesprompts: Unleashing the 3d spatial task capabilities of gpt-4v.
\newblock \emph{arXiv preprint arXiv:2312.09738}.

\bibitem[{Liu et~al.(2024{\natexlab{a}})Liu, Li, Wu, and Lee}]{liu2024visual}
Haotian Liu, Chunyuan Li, Qingyang Wu, and Yong~Jae Lee. 2024{\natexlab{a}}.
\newblock Visual instruction tuning.
\newblock \emph{Advances in neural information processing systems}, 36.

\bibitem[{Liu et~al.(2023{\natexlab{b}})Liu, Shen, Pun, and Cun}]{liu2023explicit}
Weihuang Liu, Xi~Shen, Chi-Man Pun, and Xiaodong Cun. 2023{\natexlab{b}}.
\newblock Explicit visual prompting for low-level structure segmentations.
\newblock In \emph{Proceedings of the IEEE/CVF Conference on Computer Vision and Pattern Recognition}, pages 19434--19445.

\bibitem[{Liu et~al.(2024{\natexlab{b}})Liu, Zhu, Lan, Yang, and Qiao}]{liu2024safety}
Xin Liu, Yichen Zhu, Yunshi Lan, Chao Yang, and Yu~Qiao. 2024{\natexlab{b}}.
\newblock Safety of multimodal large language models on images and text.
\newblock \emph{arXiv preprint arXiv:2402.00357}.

\bibitem[{Liu et~al.(2024{\natexlab{c}})Liu, Cai, Zhang, Yuan, and Wang}]{liu2024arondight}
Yi~Liu, Chengjun Cai, Xiaoli Zhang, Xingliang Yuan, and Cong Wang. 2024{\natexlab{c}}.
\newblock Arondight: Red teaming large vision language models with auto-generated multi-modal jailbreak prompts.
\newblock \emph{arXiv preprint arXiv:2407.15050}.

\bibitem[{Luan et~al.(2024)Luan, Feng, Chen, Wang, Zhou, and Li}]{luan2024textcot}
Bozhi Luan, Hao Feng, Hong Chen, Yonghui Wang, Wengang Zhou, and Houqiang Li. 2024.
\newblock Textcot: Zoom in for enhanced multimodal text-rich image understanding.
\newblock \emph{arXiv preprint arXiv:2404.09797}.

\bibitem[{Ma et~al.(2024{\natexlab{a}})Ma, Jiang, Wu, Yuan, and Qi}]{ma2024groma}
Chuofan Ma, Yi~Jiang, Jiannan Wu, Zehuan Yuan, and Xiaojuan Qi. 2024{\natexlab{a}}.
\newblock Groma: Localized visual tokenization for grounding multimodal large language models.
\newblock \emph{arXiv preprint arXiv:2404.13013}.

\bibitem[{Ma et~al.(2024{\natexlab{b}})Ma, Zhu, Zhang, Zhao, Wu, Huang, Hu, and Wu}]{ma2024invariant}
Huan Ma, Yan Zhu, Changqing Zhang, Peilin Zhao, Baoyuan Wu, Long-Kai Huang, Qinghua Hu, and Bingzhe Wu. 2024{\natexlab{b}}.
\newblock Invariant test-time adaptation for vision-language model generalization.
\newblock \emph{arXiv preprint arXiv:2403.00376}.

\bibitem[{Mou et~al.(2024)Mou, Wang, Xie, Wu, Zhang, Qi, and Shan}]{mou2024t2i}
Chong Mou, Xintao Wang, Liangbin Xie, Yanze Wu, Jian Zhang, Zhongang Qi, and Ying Shan. 2024.
\newblock T2i-adapter: Learning adapters to dig out more controllable ability for text-to-image diffusion models.
\newblock In \emph{Proceedings of the AAAI Conference on Artificial Intelligence}, volume~38, pages 4296--4304.

\bibitem[{Nasiriany et~al.(2024)Nasiriany, Xia, Yu, Xiao, Liang, Dasgupta, Xie, Driess, Wahid, Xu et~al.}]{nasiriany2024pivot}
Soroush Nasiriany, Fei Xia, Wenhao Yu, Ted Xiao, Jacky Liang, Ishita Dasgupta, Annie Xie, Danny Driess, Ayzaan Wahid, Zhuo Xu, et~al. 2024.
\newblock Pivot: Iterative visual prompting elicits actionable knowledge for vlms.
\newblock \emph{arXiv preprint arXiv:2402.07872}.

\bibitem[{Ni et~al.(2024)Ni, Shen, Zhang, and Zuo}]{ni2024responsible}
Minheng Ni, Yeli Shen, Lei Zhang, and Wangmeng Zuo. 2024.
\newblock Responsible visual editing.
\newblock \emph{arXiv preprint arXiv:2404.05580}.

\bibitem[{Oh et~al.(2023)Oh, Hwang, Lee, Lim, Jung, Jung, Choi, and Song}]{oh2023blackvip}
Changdae Oh, Hyeji Hwang, Hee-young Lee, YongTaek Lim, Geunyoung Jung, Jiyoung Jung, Hosik Choi, and Kyungwoo Song. 2023.
\newblock Blackvip: Black-box visual prompting for robust transfer learning.
\newblock In \emph{Proceedings of the IEEE/CVF Conference on Computer Vision and Pattern Recognition}, pages 24224--24235.

\bibitem[{Ouyang et~al.(2022)Ouyang, Wu, Jiang, Almeida, Wainwright, Mishkin, Zhang, Agarwal, Slama, Ray et~al.}]{ouyang2022training}
Long Ouyang, Jeffrey Wu, Xu~Jiang, Diogo Almeida, Carroll Wainwright, Pamela Mishkin, Chong Zhang, Sandhini Agarwal, Katarina Slama, Alex Ray, et~al. 2022.
\newblock Training language models to follow instructions with human feedback.
\newblock \emph{Advances in neural information processing systems}, 35:27730--27744.

\bibitem[{Pan et~al.(2024)Pan, Tang, Li, Fan, Chow, Yan, Chua, Zhuang, and Zhang}]{pan2024auto}
Kaihang Pan, Siliang Tang, Juncheng Li, Zhaoyu Fan, Wei Chow, Shuicheng Yan, Tat-Seng Chua, Yueting Zhuang, and Hanwang Zhang. 2024.
\newblock Auto-encoding morph-tokens for multimodal llm.
\newblock \emph{arXiv preprint arXiv:2405.01926}.

\bibitem[{Qu et~al.(2024)Qu, Li, Wang, Wang, Li, Nie, and Chua}]{qu2024unified}
Leigang Qu, Haochuan Li, Tan Wang, Wenjie Wang, Yongqi Li, Liqiang Nie, and Tat-Seng Chua. 2024.
\newblock Unified text-to-image generation and retrieval.
\newblock \emph{arXiv preprint arXiv:2406.05814}.

\bibitem[{Ren et~al.(2024)Ren, Yan, Gao, Yan, Zhang, and Li}]{ren2024you}
Huali Ren, Anli Yan, Chong-zhi Gao, Hongyang Yan, Zhenxin Zhang, and Jin Li. 2024.
\newblock Are you copying my prompt? protecting the copyright of vision prompt for vpaas via watermark.
\newblock \emph{arXiv preprint arXiv:2405.15161}.

\bibitem[{Rezaei et~al.(2024)Rezaei, Sabet, Gu, Rueckert, Torr, and Khakzar}]{rezaei2024learning}
Razieh Rezaei, Masoud~Jalili Sabet, Jindong Gu, Daniel Rueckert, Philip Torr, and Ashkan Khakzar. 2024.
\newblock Learning visual prompts for guiding the attention of vision transformers.
\newblock \emph{arXiv preprint arXiv:2406.03303}.

\bibitem[{Rombach et~al.(2022)Rombach, Blattmann, Lorenz, Esser, and Ommer}]{rombach2022high}
Robin Rombach, Andreas Blattmann, Dominik Lorenz, Patrick Esser, and Bj{\"o}rn Ommer. 2022.
\newblock High-resolution image synthesis with latent diffusion models.
\newblock In \emph{Proceedings of the IEEE/CVF conference on computer vision and pattern recognition}, pages 10684--10695.

\bibitem[{Schulhoff et~al.(2024)Schulhoff, Ilie, Balepur, Kahadze, Liu, Si, Li, Gupta, Han, Schulhoff et~al.}]{schulhoff2024prompt}
Sander Schulhoff, Michael Ilie, Nishant Balepur, Konstantine Kahadze, Amanda Liu, Chenglei Si, Yinheng Li, Aayush Gupta, HyoJung Han, Sevien Schulhoff, et~al. 2024.
\newblock The prompt report: A systematic survey of prompting techniques.
\newblock \emph{arXiv preprint arXiv:2406.06608}.

\bibitem[{Sheng et~al.(2024)Sheng, Chen, Tan, Liu, Chu, Bao, Gong, Liu, Xu, and Yu}]{sheng2024towards}
Dianmo Sheng, Dongdong Chen, Zhentao Tan, Qiankun Liu, Qi~Chu, Jianmin Bao, Tao Gong, Bin Liu, Shengwei Xu, and Nenghai Yu. 2024.
\newblock Towards more unified in-context visual understanding.
\newblock In \emph{Proceedings of the IEEE/CVF Conference on Computer Vision and Pattern Recognition}, pages 13362--13372.

\bibitem[{Shtedritski et~al.(2023)Shtedritski, Rupprecht, and Vedaldi}]{shtedritski2023does}
Aleksandar Shtedritski, Christian Rupprecht, and Andrea Vedaldi. 2023.
\newblock What does clip know about a red circle? visual prompt engineering for vlms.
\newblock In \emph{Proceedings of the IEEE/CVF International Conference on Computer Vision}, pages 11987--11997.

\bibitem[{Sun et~al.(2024)Sun, Cui, Zhang, Zhang, Yu, Wang, Rao, Liu, Huang, and Wang}]{sun2024generative}
Quan Sun, Yufeng Cui, Xiaosong Zhang, Fan Zhang, Qiying Yu, Yueze Wang, Yongming Rao, Jingjing Liu, Tiejun Huang, and Xinlong Wang. 2024.
\newblock Generative multimodal models are in-context learners.
\newblock In \emph{Proceedings of the IEEE/CVF Conference on Computer Vision and Pattern Recognition}, pages 14398--14409.

\bibitem[{Tziafas and Kasaei(2024)}]{tziafas2024towards}
Georgios Tziafas and Hamidreza Kasaei. 2024.
\newblock Towards open-world grasping with large vision-language models.
\newblock \emph{arXiv preprint arXiv:2406.18722}.

\bibitem[{Wan et~al.(2024)Wan, Cho, Stengel-Eskin, and Bansal}]{wan2024contrastive}
David Wan, Jaemin Cho, Elias Stengel-Eskin, and Mohit Bansal. 2024.
\newblock Contrastive region guidance: Improving grounding in vision-language models without training.
\newblock \emph{arXiv preprint arXiv:2403.02325}.

\bibitem[{Wang et~al.(2024{\natexlab{a}})Wang, Xu, Hu, Lan, Dong, Wang, Lee, and Lim}]{Wang2024AllIA}
Lei Wang, Wanyu Xu, Zhiqiang Hu, Yihuai Lan, Shan Dong, Hao Wang, Roy Ka-Wei Lee, and Ee-Peng Lim. 2024{\natexlab{a}}.
\newblock \href {https://api.semanticscholar.org/CorpusID:268041656} {All in an aggregated image for in-image learning}.

\bibitem[{Wang et~al.(2024{\natexlab{b}})Wang, Chen, Chen, Wu, Zhu, Zeng, Luo, Lu, Zhou, Qiao et~al.}]{wang2024visionllm}
Wenhai Wang, Zhe Chen, Xiaokang Chen, Jiannan Wu, Xizhou Zhu, Gang Zeng, Ping Luo, Tong Lu, Jie Zhou, Yu~Qiao, et~al. 2024{\natexlab{b}}.
\newblock Visionllm: Large language model is also an open-ended decoder for vision-centric tasks.
\newblock \emph{Advances in Neural Information Processing Systems}, 36.

\bibitem[{Wang et~al.(2022)Wang, Kordi, Mishra, Liu, Smith, Khashabi, and Hajishirzi}]{wang2022self}
Yizhong Wang, Yeganeh Kordi, Swaroop Mishra, Alisa Liu, Noah~A Smith, Daniel Khashabi, and Hannaneh Hajishirzi. 2022.
\newblock Self-instruct: Aligning language models with self-generated instructions.
\newblock \emph{arXiv preprint arXiv:2212.10560}.

\bibitem[{Wang et~al.(2023)Wang, Jiang, Lu, He, Chen, Wang, Zhou et~al.}]{wang2023context}
Zhendong Wang, Yifan Jiang, Yadong Lu, Pengcheng He, Weizhu Chen, Zhangyang Wang, Mingyuan Zhou, et~al. 2023.
\newblock In-context learning unlocked for diffusion models.
\newblock \emph{Advances in Neural Information Processing Systems}, 36:8542--8562.

\bibitem[{Wang et~al.(2024{\natexlab{c}})Wang, Huang, Song, Ma, and Zhang}]{wang2024promptcharm}
Zhijie Wang, Yuheng Huang, Da~Song, Lei Ma, and Tianyi Zhang. 2024{\natexlab{c}}.
\newblock Promptcharm: Text-to-image generation through multi-modal prompting and refinement.
\newblock In \emph{Proceedings of the CHI Conference on Human Factors in Computing Systems}, pages 1--21.

\bibitem[{Wu et~al.(2024{\natexlab{a}})Wu, Zhong, Xing, Lai, Liu, Wang, Chen, Zhu, Lu, Lu et~al.}]{wu2024visionllm}
Jiannan Wu, Muyan Zhong, Sen Xing, Zeqiang Lai, Zhaoyang Liu, Wenhai Wang, Zhe Chen, Xizhou Zhu, Lewei Lu, Tong Lu, et~al. 2024{\natexlab{a}}.
\newblock Visionllm v2: An end-to-end generalist multimodal large language model for hundreds of vision-language tasks.
\newblock \emph{arXiv preprint arXiv:2406.08394}.

\bibitem[{Wu et~al.(2024{\natexlab{b}})Wu, Li, Yu, Wang, Chen, Gu, Yao, Shang, and McAuley}]{wu2024commit}
Junda Wu, Xintong Li, Tong Yu, Yu~Wang, Xiang Chen, Jiuxiang Gu, Lina Yao, Jingbo Shang, and Julian McAuley. 2024{\natexlab{b}}.
\newblock Commit: Coordinated instruction tuning for multimodal large language models.
\newblock \emph{arXiv preprint arXiv:2407.20454}.

\bibitem[{Wu et~al.(2023)Wu, Wang, Zhao, Zhang, Lu, Li, and Henao}]{wu2023few}
Junda Wu, Rui Wang, Handong Zhao, Ruiyi Zhang, Chaochao Lu, Shuai Li, and Ricardo Henao. 2023.
\newblock Few-shot composition learning for image retrieval with prompt tuning.
\newblock In \emph{Proceedings of the AAAI Conference on Artificial Intelligence}, volume~37, pages 4729--4737.

\bibitem[{Wu et~al.(2024{\natexlab{c}})Wu, Jiang, Liu, Yuan, Bai, and Bai}]{wu2024general}
Junfeng Wu, Yi~Jiang, Qihao Liu, Zehuan Yuan, Xiang Bai, and Song Bai. 2024{\natexlab{c}}.
\newblock General object foundation model for images and videos at scale.
\newblock In \emph{Proceedings of the IEEE/CVF Conference on Computer Vision and Pattern Recognition}, pages 3783--3795.

\bibitem[{Wu et~al.(2024{\natexlab{d}})Wu, Cai, Ji, Li, Huang, Luo, Fei, Sun, and Ji}]{wu2024controlmllm}
Mingrui Wu, Xinyue Cai, Jiayi Ji, Jiale Li, Oucheng Huang, Gen Luo, Hao Fei, Xiaoshuai Sun, and Rongrong Ji. 2024{\natexlab{d}}.
\newblock Controlmllm: Training-free visual prompt learning for multimodal large language models.
\newblock \emph{arXiv preprint arXiv:2407.21534}.

\bibitem[{Wu et~al.(2024{\natexlab{e}})Wu, Huang, and Wang}]{wu2024dora}
Tung-Yu Wu, Sheng-Yu Huang, and Yu-Chiang~Frank Wang. 2024{\natexlab{e}}.
\newblock Dora: 3d visual grounding with order-aware referring.
\newblock \emph{arXiv preprint arXiv:2403.16539}.

\bibitem[{Wu et~al.(2024{\natexlab{f}})Wu, Wang, Tang, Wu, He, Ouyang, Wu, and Torr}]{wu2024dettoolchain}
Yixuan Wu, Yizhou Wang, Shixiang Tang, Wenhao Wu, Tong He, Wanli Ouyang, Jian Wu, and Philip Torr. 2024{\natexlab{f}}.
\newblock Dettoolchain: A new prompting paradigm to unleash detection ability of mllm.
\newblock \emph{arXiv preprint arXiv:2403.12488}.

\bibitem[{Xie et~al.(2021)Xie, Wang, Yu, Anandkumar, Alvarez, and Luo}]{xie2021segformer}
Enze Xie, Wenhai Wang, Zhiding Yu, Anima Anandkumar, Jose~M Alvarez, and Ping Luo. 2021.
\newblock Segformer: Simple and efficient design for semantic segmentation with transformers.
\newblock \emph{Advances in neural information processing systems}, 34:12077--12090.

\bibitem[{Xie et~al.(2024)Xie, Han, Bain, Nagrani, Varol, Xie, and Zisserman}]{xie2024autoad}
Junyu Xie, Tengda Han, Max Bain, Arsha Nagrani, G{\"u}l Varol, Weidi Xie, and Andrew Zisserman. 2024.
\newblock Autoad-zero: A training-free framework for zero-shot audio description.
\newblock \emph{arXiv preprint arXiv:2407.15850}.

\bibitem[{Xu et~al.(2024{\natexlab{a}})Xu, Liu, Wang, and Fu}]{xu2024towards}
Chengming Xu, Chen Liu, Yikai Wang, and Yanwei Fu. 2024{\natexlab{a}}.
\newblock Towards global optimal visual in-context learning prompt selection.
\newblock \emph{arXiv preprint arXiv:2405.15279}.

\bibitem[{Xu et~al.(2024{\natexlab{b}})Xu, Liu, Pasupat, Kazemi et~al.}]{xu2024context}
Xin Xu, Yue Liu, Panupong Pasupat, Mehran Kazemi, et~al. 2024{\natexlab{b}}.
\newblock In-context learning with retrieved demonstrations for language models: A survey.
\newblock \emph{arXiv preprint arXiv:2401.11624}.

\bibitem[{Yan et~al.(2024)Yan, Yang, Wu, Zhu, Yang, Li, Lin, Wang, McAuley, Gao et~al.}]{yan2024list}
An~Yan, Zhengyuan Yang, Junda Wu, Wanrong Zhu, Jianwei Yang, Linjie Li, Kevin Lin, Jianfeng Wang, Julian McAuley, Jianfeng Gao, et~al. 2024.
\newblock List items one by one: A new data source and learning paradigm for multimodal llms.
\newblock \emph{arXiv preprint arXiv:2404.16375}.

\bibitem[{Yang et~al.(2023)Yang, Zhang, Li, Zou, Li, and Gao}]{yang2023set}
Jianwei Yang, Hao Zhang, Feng Li, Xueyan Zou, Chunyuan Li, and Jianfeng Gao. 2023.
\newblock Set-of-mark prompting unleashes extraordinary visual grounding in gpt-4v.
\newblock \emph{arXiv preprint arXiv:2310.11441}.

\bibitem[{Yang et~al.(2024{\natexlab{a}})Yang, Wang, Li, Wang, and Yang}]{yang2024fine}
Lingfeng Yang, Yueze Wang, Xiang Li, Xinlong Wang, and Jian Yang. 2024{\natexlab{a}}.
\newblock Fine-grained visual prompting.
\newblock \emph{Advances in Neural Information Processing Systems}, 36.

\bibitem[{Yang et~al.(2024{\natexlab{b}})Yang, Peng, Shen, Yang, Hu, Qiu, Koike et~al.}]{yang2024imagebrush}
Yifan Yang, Houwen Peng, Yifei Shen, Yuqing Yang, Han Hu, Lili Qiu, Hideki Koike, et~al. 2024{\natexlab{b}}.
\newblock Imagebrush: Learning visual in-context instructions for exemplar-based image manipulation.
\newblock \emph{Advances in Neural Information Processing Systems}, 36.

\bibitem[{Ye et~al.(2024)Ye, Zheng, Ma, Cao, Lai, Rehg, and Zhang}]{ye2024mm}
Wenqian Ye, Guangtao Zheng, Yunsheng Ma, Xu~Cao, Bolin Lai, James~M Rehg, and Aidong Zhang. 2024.
\newblock Mm-spubench: Towards better understanding of spurious biases in multimodal llms.
\newblock \emph{arXiv preprint arXiv:2406.17126}.

\bibitem[{Ying et~al.(2024)Ying, Liu, Zhang, Yu, Liang, Liu, and Tao}]{ying2024jailbreak}
Zonghao Ying, Aishan Liu, Tianyuan Zhang, Zhengmin Yu, Siyuan Liang, Xianglong Liu, and Dacheng Tao. 2024.
\newblock Jailbreak vision language models via bi-modal adversarial prompt.
\newblock \emph{arXiv preprint arXiv:2406.04031}.

\bibitem[{Yoon et~al.(2024)Yoon, Yu, and Bansal}]{yoon2024raccoon}
Jaehong Yoon, Shoubin Yu, and Mohit Bansal. 2024.
\newblock Raccoon: Remove, add, and change video content with auto-generated narratives.
\newblock \emph{arXiv preprint arXiv:2405.18406}.

\bibitem[{Zhang et~al.(2024{\natexlab{a}})Zhang, Fei, Yao, Ji, Li, Liu, and Chua}]{zhang2024vpgtrans}
Ao~Zhang, Hao Fei, Yuan Yao, Wei Ji, Li~Li, Zhiyuan Liu, and Tat-Seng Chua. 2024{\natexlab{a}}.
\newblock Vpgtrans: Transfer visual prompt generator across llms.
\newblock \emph{Advances in Neural Information Processing Systems}, 36.

\bibitem[{Zhang et~al.(2023{\natexlab{a}})Zhang, Puspitasari, Zheng, Li, Qiao, Kang, Shan, Zhang, Qin, Rameau et~al.}]{zhang2023survey}
Chaoning Zhang, Fachrina~Dewi Puspitasari, Sheng Zheng, Chenghao Li, Yu~Qiao, Taegoo Kang, Xinru Shan, Chenshuang Zhang, Caiyan Qin, Francois Rameau, et~al. 2023{\natexlab{a}}.
\newblock A survey on segment anything model (sam): Vision foundation model meets prompt engineering.
\newblock \emph{arXiv preprint arXiv:2306.06211}.

\bibitem[{Zhang et~al.(2023{\natexlab{b}})Zhang, Li, Zou, Liu, Li, Yang, and Zhang}]{zhang2023simple}
Hao Zhang, Feng Li, Xueyan Zou, Shilong Liu, Chunyuan Li, Jianwei Yang, and Lei Zhang. 2023{\natexlab{b}}.
\newblock A simple framework for open-vocabulary segmentation and detection.
\newblock In \emph{Proceedings of the IEEE/CVF International Conference on Computer Vision}, pages 1020--1031.

\bibitem[{Zhang et~al.(2024{\natexlab{b}})Zhang, Huang, Jin, and Lu}]{zhang2024vision}
Jingyi Zhang, Jiaxing Huang, Sheng Jin, and Shijian Lu. 2024{\natexlab{b}}.
\newblock Vision-language models for vision tasks: A survey.
\newblock \emph{IEEE Transactions on Pattern Analysis and Machine Intelligence}.

\bibitem[{Zhang et~al.(2024{\natexlab{c}})Zhang, Zhao, Ying, Ma, and Lee}]{zhang2024omagent}
Lu~Zhang, Tiancheng Zhao, Heting Ying, Yibo Ma, and Kyusong Lee. 2024{\natexlab{c}}.
\newblock Omagent: A multi-modal agent framework for complex video understanding with task divide-and-conquer.
\newblock \emph{arXiv preprint arXiv:2406.16620}.

\bibitem[{Zhang et~al.(2023{\natexlab{c}})Zhang, Rao, and Agrawala}]{zhang2023adding}
Lvmin Zhang, Anyi Rao, and Maneesh Agrawala. 2023{\natexlab{c}}.
\newblock Adding conditional control to text-to-image diffusion models.
\newblock In \emph{Proceedings of the IEEE/CVF International Conference on Computer Vision}, pages 3836--3847.

\bibitem[{Zhang et~al.(2024{\natexlab{d}})Zhang, Huang, Deng, Tang, Ouyang, He, and Zhang}]{zhang2024agent3d}
Sha Zhang, Di~Huang, Jiajun Deng, Shixiang Tang, Wanli Ouyang, Tong He, and Yanyong Zhang. 2024{\natexlab{d}}.
\newblock Agent3d-zero: An agent for zero-shot 3d understanding.
\newblock \emph{arXiv preprint arXiv:2403.11835}.

\bibitem[{Zhang et~al.(2024{\natexlab{e}})Zhang, Li, Fei, Yuan, Wu, Ji, Loy, and Yan}]{zhang2024omg}
Tao Zhang, Xiangtai Li, Hao Fei, Haobo Yuan, Shengqiong Wu, Shunping Ji, Chen~Change Loy, and Shuicheng Yan. 2024{\natexlab{e}}.
\newblock Omg-llava: Bridging image-level, object-level, pixel-level reasoning and understanding.
\newblock \emph{arXiv preprint arXiv:2406.19389}.

\bibitem[{Zhang et~al.(2024{\natexlab{f}})Zhang, Cai, Zhang, Zhuang, and Mao}]{zhang2024earthmarker}
Wei Zhang, Miaoxin Cai, Tong Zhang, Yin Zhuang, and Xuerui Mao. 2024{\natexlab{f}}.
\newblock Earthmarker: A visual prompt learning framework for region-level and point-level remote sensing imagery comprehension.
\newblock \emph{arXiv preprint arXiv:2407.13596}.

\bibitem[{Zhang et~al.(2024{\natexlab{g}})Zhang, Dong, Zhang, Min, Su, and Zhu}]{zhang2024exploring}
Yichi Zhang, Yinpeng Dong, Siyuan Zhang, Tianzan Min, Hang Su, and Jun Zhu. 2024{\natexlab{g}}.
\newblock Exploring the transferability of visual prompting for multimodal large language models.
\newblock In \emph{Proceedings of the IEEE/CVF Conference on Computer Vision and Pattern Recognition}, pages 26562--26572.

\bibitem[{Zhang et~al.(2024{\natexlab{h}})Zhang, Ma, Zhang, and Bai}]{zhang2024psalm}
Zheng Zhang, Yeyao Ma, Enming Zhang, and Xiang Bai. 2024{\natexlab{h}}.
\newblock Psalm: Pixelwise segmentation with large multi-modal model.
\newblock \emph{arXiv preprint arXiv:2403.14598}.

\bibitem[{Zhong et~al.(2024)Zhong, Wu, Li, Barton, Du, Sam, Bouyarmane, Tutar, and Huang}]{zhong2024enhancing}
Wenliang Zhong, Wenyi Wu, Qi~Li, Rob Barton, Boxin Du, Shioulin Sam, Karim Bouyarmane, Ismail Tutar, and Junzhou Huang. 2024.
\newblock Enhancing multimodal large language models with multi-instance visual prompt generator for visual representation enrichment.
\newblock \emph{arXiv preprint arXiv:2406.02987}.

\bibitem[{Zhou et~al.(2024{\natexlab{a}})Zhou, Qin, Yin, Huang, Zhang, Sheng, Qiao, and Shao}]{zhou2024minedreamer}
Enshen Zhou, Yiran Qin, Zhenfei Yin, Yuzhou Huang, Ruimao Zhang, Lu~Sheng, Yu~Qiao, and Jing Shao. 2024{\natexlab{a}}.
\newblock Minedreamer: Learning to follow instructions via chain-of-imagination for simulated-world control.
\newblock \emph{arXiv preprint arXiv:2403.12037}.

\bibitem[{Zhou et~al.(2024{\natexlab{b}})Zhou, Zhou, Hu, Lu, Gao, and Zhang}]{zhou2024image}
Qiji Zhou, Ruochen Zhou, Zike Hu, Panzhong Lu, Siyang Gao, and Yue Zhang. 2024{\natexlab{b}}.
\newblock Image-of-thought prompting for visual reasoning refinement in multimodal large language models.
\newblock \emph{arXiv preprint arXiv:2405.13872}.

\end{thebibliography}

\end{document}